\documentclass{article}

\usepackage{microtype}
\usepackage{graphicx}
\usepackage{subcaption}
\usepackage{booktabs} 

\usepackage{hyperref}
\usepackage{url}            
\usepackage{amsfonts}       
\usepackage{nicefrac}       
\usepackage{microtype}      
\usepackage{xcolor}         
\usepackage{enumitem}
\usepackage{multirow}
\usepackage{wrapfig}
\usepackage{adjustbox}
\usepackage{amsmath}
\usepackage{float}

\usepackage[preprint]{icml2026}

\usepackage{amsmath}
\usepackage{amssymb}
\usepackage{mathtools}
\usepackage{amsthm}

\usepackage[capitalize,noabbrev]{cleveref}

\theoremstyle{plain}

\theoremstyle{definition}

\theoremstyle{remark}

\usepackage[textsize=tiny]{todonotes}

\icmltitlerunning{Spatially-Adaptive Gradient Re-parameterization for 3D Large Kernel Optimization}

\begin{document}
\twocolumn[
  \icmltitle{Spatially-Adaptive Gradient Re-parameterization for 3D Large Kernel Optimization}

  \icmlsetsymbol{equal}{*}

  \begin{icmlauthorlist}
    \icmlauthor{Ho Hin Lee}{yyy}
    \icmlauthor{Quan Liu}{comp}
    \icmlauthor{Shunxing Bao}{yyy}
    \icmlauthor{Yuankai Huo}{yyy}
    \icmlauthor{Bennett A. Landman}{yyy}
  \end{icmlauthorlist}

  \icmlaffiliation{yyy}{Department of Electrical and Computer Engineering, Vanderbilt University, Nashville, USA}
  \icmlaffiliation{comp}{Accenture, USA}

  \icmlcorrespondingauthor{Ho Hin Lee}{ho.hin.lee@vanderbilt.edu}

  \icmlkeywords{Large Kernel Convolution, Weight Re-parameterization, Effective Receptive Field Modeling, Medical Imaging}

  \vskip 0.3in
]

\printAffiliationsAndNotice{} 

\begin{abstract}
Large kernel convolutions offer a scalable alternative to vision transformers for high-resolution 3D volumetric analysis, yet naively increasing kernel size often leads to optimization instability. Motivated by the spatial bias inherent in effective receptive fields (ERFs), we theoretically demonstrate that structurally re-parameterized blocks induce spatially varying learning rates that are crucial for convergence. Leveraging this insight, we introduce Rep3D, a framework that employs a lightweight modulation network to generate receptive-biased scaling masks, adaptively re-weighting kernel updates within a plain encoder architecture. This approach unifies spatial inductive bias with optimization-aware learning, avoiding the complexity of multi-branch designs while ensuring robust local-to-global convergence. Extensive evaluations on five 3D segmentation benchmarks demonstrate that Rep3D consistently outperforms state-of-the-art transformer and fixed-prior baselines. The source code is publicly available at \url{https://github.com/leeh43/Rep3D}.
\end{abstract}

\section{Introduction}
The landscape of medical vision models has evolved rapidly, expanding from early convolutional architectures to modern transformer-based designs. In particular, Vision Transformers (ViTs) have gained traction for their ability to model long-range dependencies using multi-head self-attention and minimal inductive bias \cite{dosovitskiy2020image}. In parallel, the community has revisited large kernel convolutions as a scalable alternative to attention mechanisms, particularly in the context of high-resolution 3D volumetric data \cite{liu2022convnet, lee20223d}. Despite architectural differences, both ViTs and large-kernel CNNs share a central goal: expanding the effective receptive fields (ERFs) to enable rich spatial context aggregation. However, simply increasing kernel size does not guarantee improved performance. Prior work has shown that naive enlargement of convolutional filters can result in saturated or degraded accuracy across various segmentation tasks \cite{ding2022scaling, lee2023scaling}. Unlike ViTs, which adaptively attend to spatial content, standard convolutions rely on static, weight-shared kernels and lack the ability to modulate importance across spatial positions. This limitation prompts our first research question:
\textbf{Can we incorporate spatial priors into large kernel convolutions to improve learning effectiveness?}

Recent advances in structural re-parameterization offer a promising direction. Methods such as RepLKNet \cite{ding2022scaling}, SLaK \cite{liu2022more}, and PELK \cite{chen2024pelk} scale kernels to extreme sizes (e.g., $31\times31$, $51\times51$, $101\times101$) by combining parallel branches of ``large + small'' convolutions into what is referred to as a Constant-Scale Linear Addition (CSLA) block. These parallel paths are merged into a single kernel at inference time, enabling efficient deployment while capturing multi-scale features during training. Interestingly, we observe that CSLA blocks naturally encode spatial learning bias: elements near the kernel center tend to converge faster than those on the periphery. This mirrors diffusion-like gradient propagation in ERFs starting from the center and expanding outward. These observations suggest that convergence dynamics are not uniform across the kernel, but instead spatially structured. This leads to our second question:
\textbf{Can we explicitly model this diffusion pattern as a learnable spatial prior to re-weight kernel element updates during training?}

To address this, we first provide a theoretical analysis of the optimization dynamics in CSLA-based re-parameterized convolutions. We show that each branch (e.g., small vs. large kernels) implicitly operates under a distinct learning rate, leading to element-wise differences in convergence speed. These dynamics correlate with ERF visualizations and share characteristics with spatial frequency patterns in human visual perception \cite{kulikowski1982theory}. Inspired by this, we propose a novel receptive bias re-parameterization strategy that encodes spatial distance from the kernel center as a spatial bias prior on learning convergence. We implement this as a low-rank modulation mechanism that generates spatial scaling factors for kernel weights, allowing the optimizer to emphasize local versus global regions adaptively for gradient back-propagation. Building on this insight, we present Rep3D, a 3D convolutional architecture that integrates large kernel convolutions (e.g., $21\times21\times21$) with our proposed re-parameterization approach. Unlike prior approaches that rely on multi-branch structures, Rep3D employs a plain and efficient encoder to reduce complexity while preserving representational capacity. We evaluate Rep3D across five challenging volumetric medical segmentation benchmarks and show that it consistently outperforms state-of-the-art transformer- and CNN-based models. Our key contributions are as follows:
\begin{itemize}
    \item We propose Rep3D, a 3D CNN with large kernel convolutions and a streamlined encoder design that achieves state-of-the-art (SOTA) performance on multi-scale (i.e. from organs/tissues to tumors) segmentation benchmarks.
    \item We propose a novel and theoretically grounded re-parameterization approach that models ERF diffusion as a learnable spatial bias prior, enabling element-wise modulation of gradient convergence for training.
    \item We validate our method on five challenging 3D medical imaging benchmarks under direct training settings, achieving consistent and significant improvements across all datasets.
\end{itemize}

\begin{figure*}[t!]
\centering
\includegraphics[width=\textwidth]{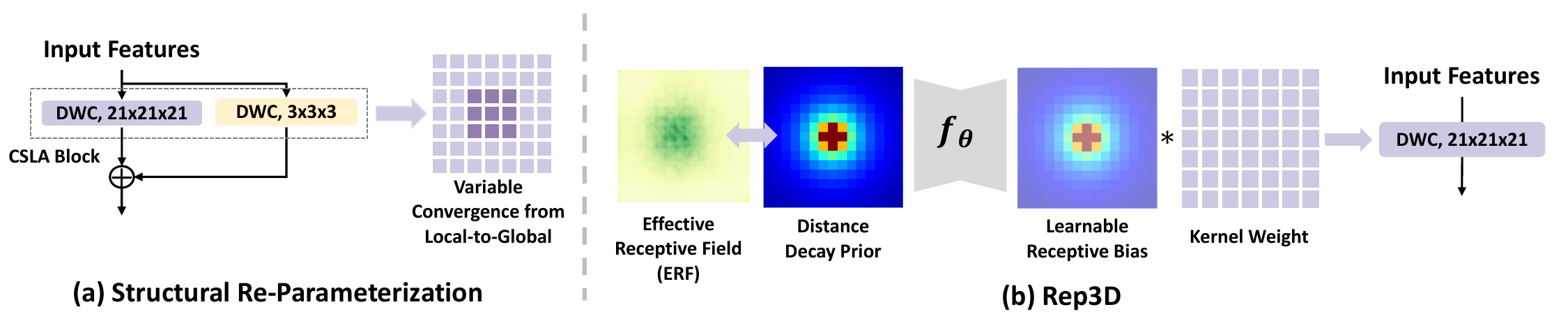}
\caption{(a) Traditional structural re-parameterization methods (e.g., CSLA blocks) re-parameterize small and large kernel convolutions to improve representational capacity, but apply linear optimization with the same learning rate across the kernels, demonstrating faster convergence in local regions than global ones. (b) In contrast, Rep3D introduces a learnable spatial bias via a generator network $f_\theta$, which modulates each element in the large kernel using a prior based on distance decay. This adaptive modulation enables local-to-global update dynamics aligned with ERF behavior, enhancing both training stability and model performance for 3D volumetric tasks.} \label{fig:idea_figure}
\end{figure*}

\section{Related Work}
\textbf{CNN-based 3D Models:} Foundational architectures such as 3D U-Net \cite{cciccek20163d}, V-Net \cite{milletari2016v}, and nnU-Net \cite{isensee2021nnu} have played a pivotal role in establishing the standard for volumetric medical image segmentation. These models rely on encoder-decoder designs with dense skip connections, offering a strong balance between spatial resolution and semantic representation. Due to their stability, interpretability, and effectiveness without requiring large-scale pre-training, they remain widely used in clinical and research benchmarks. However, their inherently local receptive fields limit their ability to capture long-range dependencies, motivating subsequent architectural innovations aimed at expanding the effective receptive field (ERF).

\textbf{Vision Transformer-based and Hybrid Models:} To overcome the locality constraints of CNNs, transformer-based architectures such as UNETR \cite{hatamizadeh2022unetr} and SwinUNETR \cite{hatamizadeh2022swin} introduce global self-attention mechanisms that model distant spatial dependencies more effectively (e.g., follow-up models like nnFormer \cite{zhou2021nnformer}, Swin-Unet \cite{cao2021swin}, and SwinBTS \cite{jiang2022swinbts}). These models encode context across entire volumes through hierarchical token representations, marking a major shift in design philosophy. However, they typically require large-scale pre-training and introduce significant computational complexity due to the quadratic scaling of attention, particularly problematic in 3D volumetric settings. Additionally, their reliance on patch-based tokenization can compromise fine-grained spatial precision—crucial in dense prediction tasks like medical segmentation. 

\textbf{Large Kernel Convolution Networks:} A more recent and efficient alternative to transformers involves expanding the ERF through large kernel convolutions, as demonstrated by models such as ConvNeXt \cite{liu2022convnet}, 3D UX-Net \cite{lee20223d}, and MedNeXt \cite{roy2023mednext}. These architectures leverage depth-wise or separable convolutions to approximate global context modeling while preserving the simplicity and inductive biases of convolutional designs. However, studies in 2D vision backbones (e.g., RepLKNet \cite{ding2022scaling} (kernel size: $31\times31$), SLaK \cite{liu2022more} (kernel size: $51\times51$)) reveal that naively scaling up kernel size leads to saturation or performance degradation in the absence of additional structural guidance. This key insight motivates the design of Rep3D, which augments large 3D kernels with a learnable spatial prior inspired by ERF theory. By explicitly guiding convergence dynamics across kernel elements, Rep3D enables more effective utilization of large kernels, bridging the gap between CNN efficiency and transformer-like contextual modeling.

\textbf{The Integration of Weight Re-parameterization.} Structural re-parameterization (SR) has emerged as a powerful paradigm to enhance CNN training without altering inference-time complexity. Models like RepVGG \cite{ding2021repvgg} and OREPA \cite{hu2022online} employ additional convolution branches (e.g., $1\times1$ or identity paths) during training to improve gradient flow and feature diversity. These branches are merged into a single convolution kernel post-training, allowing for efficient inference. RepLKNet \cite{ding2022scaling} and SLaK \cite{liu2022more} extend this approach to large 2D kernels, increasing the receptive field while maintaining tractable inference cost via kernel decomposition or sparse groups. A complementary line of work focuses on gradient re-parameterization instead of modifying model weights directly. RepOptimizer \cite{ding2022re}, for example, modifies the back-propagation process by applying learnable scaling to gradient updates, enabling effective training of plain CNNs. While much of the re-parameterization research has focused on 2D natural images, extending these methods to 3D medical imaging presents unique challenges. Volumetric kernels require significantly more parameters, and naïve kernel expansion leads to high computational costs and optimization instability. 3D RepUX-Net \cite{lee2023scaling} demonstrated the initial attempt of adapting weight re-parameterization to 3D medical imaging and scale large depthwise kernels with fixed prior context, but it still lacks flexibility in adapting dynamic variations in fine-grained semantics for learning convergence. To bridge this gap, there is growing interest in using spatial priors or effective receptive field modeling to guide re-parameterization for large kernel learning in the 3D setting.

\section{Rep3D}
\begin{figure*}[t!]
\centering
\includegraphics[width=\textwidth]{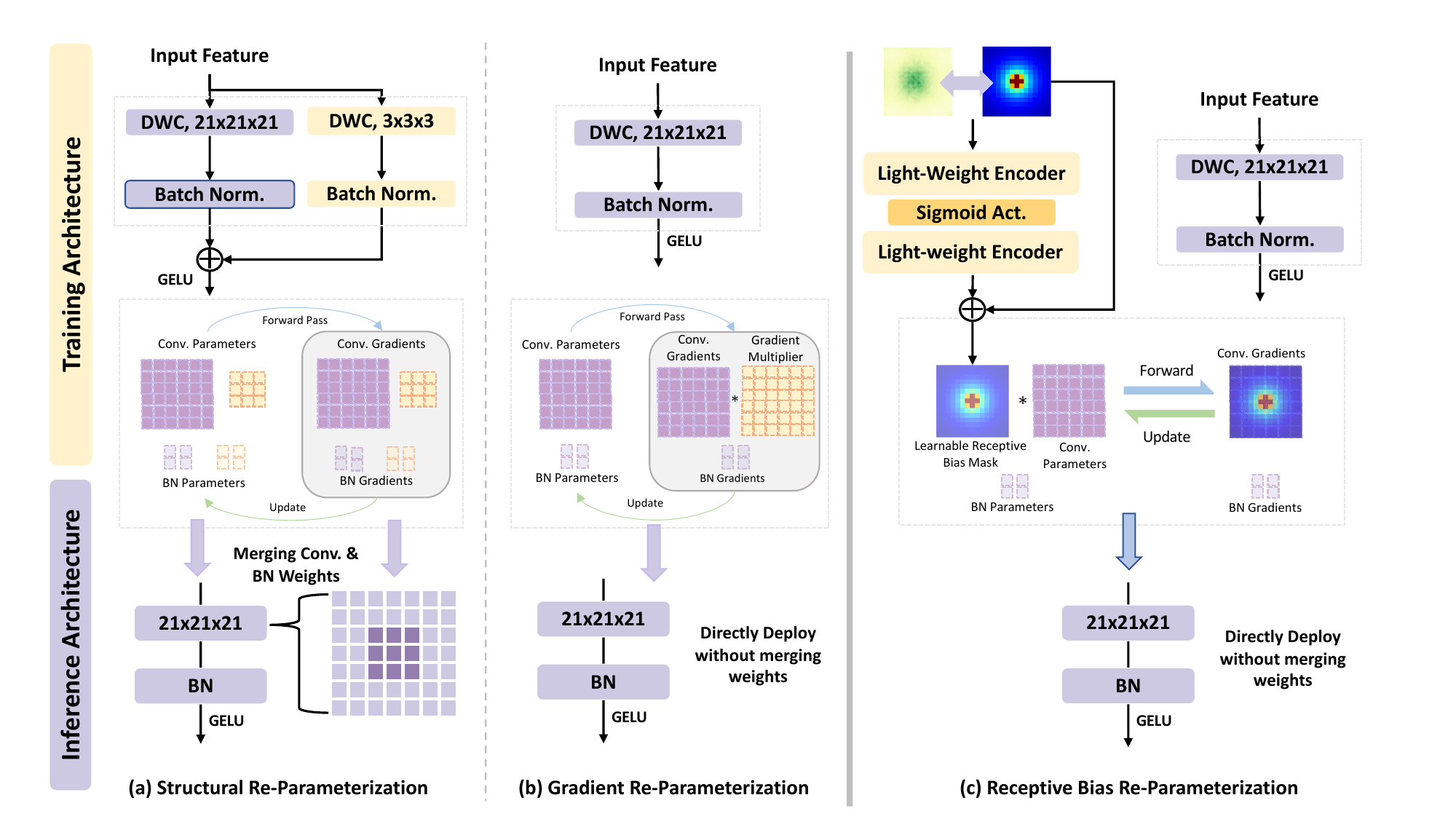}
\caption{In contrast to (a) structural or (b) gradient-based re-parameterization, Rep3D introduces a novel re-parameterization strategy that injects a learnable spatial bias into large kernel convolutions for optimization. During training, a lightweight generator network produces a modulation mask conditioned on a distance-based prior, which adaptively scales gradient updates across the kernel. This enables spatially-aware learning dynamics that reflect local-to-global variations in the effective receptive field (ERF).} \label{fig:method_figure}
\end{figure*}

Rep3D rethinks the training dynamics of large-kernel convolution by explicitly embedding spatial bias, derived from effective receptive fields (ERFs), into the optimization process. Motivated by structural re-parameterization (SR) and the distinctive gradient behavior observed in ERFs, Rep3D introduces a low-rank, learnable re-parameterization that adapts element-wise update behavior across the kernel. We first derive the theoretical equivalence between parallel convolution branches and their single-operator counterparts, showing that a ``large + small'' convolution block (as in RepLKNet~\cite{ding2022scaling}) implicitly assigns spatially varying learning rates. We then translate this insight into a unified formulation and construct a lightweight generator that outputs a convergence-aware modulation mask, as shown in Figure \ref{fig:method_figure}. The output modulated mask models fine-grained learning dynamics during training, improving both scalability and performance in 3D tasks with large kernel convolution.

\subsection{Variable Learning Convergence in Parallel Branch}
As shown in Figure~\ref{fig:method_figure}, the learning convergence of large kernel convolution can be improved by either adding up the encoded outputs of parallel branches weighted by diverse scales with SR (RepLKNet \cite{ding2022scaling}) or performing Gradient Re-parameterization (GR) by multiplying with constant values (RepOptimizer \cite{ding2022re}) in a Single Operator (SO). Inspired by the concepts of structural re-parameterization (SR) and gradient re-parameterization, we extend the theoretical derivation from RepOptimizer and observe the variable learning rate across branches. We begin by analyzing the CSLA block, a basic two-branch design used in SR-based networks. Let $X$ denote the input feature map, and let $W_L, W_S$ be large and small 3D convolution kernels, scaled by fixed positive scalars $\alpha_L$ and $\alpha_S$, respectively. The output of the CSLA module is:
\begin{equation}
    Y_{\text{CSLA}} = \alpha_L (X \ast W_L) + \alpha_S (X \ast W_S),
    \label{eq:csla_output}
\end{equation}
where $\ast$ denotes 3D convolution. To unify the branches into a single equivalent convolution for efficient inference, we define a single-operator (SO) form:
\begin{equation}
    Y_{\text{SO}} = X \ast W',
\end{equation}
where the equivalent kernel $W'$ is a linear combination of the two branches:
\begin{equation}
    W' = \alpha_L W_L + \alpha_S W_S.
\end{equation}
During training with first-order optimization (i.e., SGD, AdamW) and step size $\lambda$, we apply the stochastic gradient descent rule and update the gradients for the parallel branches as follows:
\begin{equation}
    W'_{t+1} = W'_t - \lambda \frac{\partial \mathcal{L}}{\partial W'_t}.
\end{equation}
As the parallel branch architecture updates $W_L$ and $W_S$ independently, the specific updates are:
\begin{align}
    W_{L(t+1)} &= W_{L(t)} - \lambda_L \frac{\partial \mathcal{L}}{\partial W_{L(t)}}, \\ 
    W_{S(t+1)} &= W_{S(t)} - \lambda_S \frac{\partial \mathcal{L}}{\partial W_{S(t)}},
\end{align}
where $\lambda_L$ and $\lambda_S$ are the learning rates for the corresponding branches. Substituting these into the equivalent kernel formulation yields:
\begin{align}
    W'_{t+1} &= \alpha_L W_{L(t+1)} + \alpha_S W_{S(t+1)} \nonumber \\
    &= \alpha_L \left(W_{L(t)} - \lambda_L \frac{\partial \mathcal{L}}{\partial W_{L(t)}}\right) \nonumber \\
    &\quad + \alpha_S \left(W_{S(t)} - \lambda_S \frac{\partial \mathcal{L}}{\partial W_{S(t)}}\right) \nonumber \\
    &= W'_t - \lambda_L \alpha_L \frac{\partial \mathcal{L}}{\partial W_{L(t)}} - \lambda_S \alpha_S \frac{\partial \mathcal{L}}{\partial W_{S(t)}}.
    \label{eq:combined_update}
\end{align}
From \eqref{eq:combined_update}, we observe that each branch can be optimized differently with distinct learning rates toward each kernel, leading to two distinctive scenarios:
\begin{equation}
    W'_{t+1}=
    \begin{cases}
      \begin{aligned}
        &W'_t - \lambda \Big(\alpha_L \frac{\partial \mathcal{L}}{\partial W_{L(t)}} \\
        &\quad + \alpha_S \frac{\partial \mathcal{L}}{\partial W_{S(t)}}\Big),
      \end{aligned} & \text{if } \lambda_L = \lambda_S, \\[18pt]
      \begin{aligned}
        &W'_t - \lambda_L \alpha_L \frac{\partial \mathcal{L}}{\partial W_{L(t)}} \\
        &\quad - \lambda_S \alpha_S \frac{\partial \mathcal{L}}{\partial W_{S(t)}},
      \end{aligned} & \text{if } \lambda_L \neq \lambda_S.
    \end{cases}
    \label{eq:update_scenarios}
\end{equation}
By the chain rule, we further derive:
\begin{align}
    \frac{\partial \mathcal{L}}{\partial W_{L(t)}} &= \frac{\partial \mathcal{L}}{\partial Y_{\text{CSLA}}} \cdot \frac{\partial Y_{\text{CSLA}}}{\partial W_{L(t)}} \nonumber \\
    &= \alpha_L \cdot \frac{\partial \mathcal{L}}{\partial Y_{\text{CSLA}}} \cdot \frac{\partial (X \ast W_L)}{\partial W_{L(t)}}, \label{eq:grad_L} \\
    \frac{\partial \mathcal{L}}{\partial W_{S(t)}} &= \frac{\partial \mathcal{L}}{\partial Y_{\text{CSLA}}} \cdot \frac{\partial Y_{\text{CSLA}}}{\partial W_{S(t)}} \nonumber \\
    &= \alpha_S \cdot \frac{\partial \mathcal{L}}{\partial Y_{\text{CSLA}}} \cdot \frac{\partial (X \ast W_S)}{\partial W_{S(t)}}. \label{eq:grad_S}
\end{align}
To validate the above derivation, we perform ablation studies and find that variable learning rates for each branch (i.e., $\lambda_S=0.0006, \lambda_L=0.0002$) demonstrate the best performance with stochastic gradient descent. Since $W_S$ has a smaller receptive field than $W_L$, and $W_S$ primarily contributes to the central region of the equivalent kernel $W'$, we argue the following:
\begin{itemize}
    \item \textbf{Central region of $W'$:} receives gradient contributions from both $W_L$ and $W_S$, resulting in faster convergence and stronger local learning.
    \item \textbf{Peripheral region of $W'$:} receives gradients only from $W_L$, leading to slower convergence but maintaining global contextual awareness.
\end{itemize}
Because the coefficients $\alpha_L$ and $\alpha_S$ modulate spatially distinct regions (large kernel contributions dominate the periphery, small kernel contributions dominate the central region), the two-branch block demonstrates an effective learning-rate field of:
\begin{equation}
    \lambda_{\mathrm{eff}}(\Delta x)=
    \begin{cases}
      \alpha_L\,\lambda_L, & \text{peripheral offsets},\\[4pt]
      \alpha_L\,\lambda_L+\alpha_S\,\lambda_S, & \text{central offsets},
    \end{cases}
    \label{eq:lambda_eff}
\end{equation}
where $\lambda_{\mathrm{eff}}$ is the effective element-wise learning rate inherited from the two branch-specific updates.

\subsection{Low-Rank Receptive Bias Modeling (LRBM)}
As the above theory validates the correlation between variable learning and local-to-global gradient dynamics in ERF, we argue that such receptive bias can enhance the efficiency of learning large convolution kernels. We model the diffusion behavior of ERF with a reciprocal distance decay function $f_d$ and generate a prior mapping $P \in \mathbb{R}^{C \times 1 \times K \times K \times K}$ for weight re-parameterization as follows:
\begin{align}
    f_d(x, y, z, c) &= \sqrt {(x- c)^2 + (y - c)^2 + (z - c)^2}, \\
    P &= \frac{\beta}{f_d(x_k, y_k, z_k, c) + \beta},
\end{align}
where $k$ and $c$ are the element and central index of the kernel weight, and $\beta$ is a learnable parameter initialized to 0 that controls the weight distribution of the distance mapping. However, such a fixed prior mapping lacks the flexibility to adapt the weighting importance dynamically across the fine-grained semantic variations in medical imaging. To address this, we propose to adapt learnable spatial bias by co-training a lightweight 2-layer generator network $f_\theta: \mathbb{R}^{C \times 1 \times K \times K \times K} \rightarrow \mathbb{R}^{C \times 1 \times K \times K \times K}$. We generate an adaptive mask $M$ for depthwise convolution kernels with low computation cost as follows:
\begin{equation}
    M = P + f_\theta(P),
\end{equation}
with the generator defined as:
\begin{equation}
    f_\theta(P) = \mathrm{Norm}_2\Big(\mathrm{DConv}_2\big( \sigma( \mathrm{Norm}_1(\mathrm{DConv}_1(P))) \big)\Big),
\end{equation}
where $\mathrm{DConv}_1$ and $\mathrm{DConv}_2$ are 3D depthwise convolutions with kernel size 7 and padding 3, $\mathrm{Norm}_1$ and $\mathrm{Norm}_2$ are layer normalizations, and $\sigma$ is a sigmoid activation ensuring all scaling values are between 0 and 1. Such a learnable function aims to capture the dynamic weighting of each kernel element across local to global contexts while preserving computational efficiency. The resulting modulation mask $M$ is then used to re-parameterize the kernel weights:
\begin{equation}
    W_{\mathrm{eff}} = W \odot M,
\end{equation}
where $W$ is the original convolution kernel and $\odot$ denotes element-wise multiplication. Importantly, the mask is applied during training only, and the learned generator can be removed during inference for efficiency.

\subsection{Network Architecture}
The overall network architecture to validate Rep3D builds upon the encoder–decoder structure of 3D UX-Net~\cite{lee20223d}, which processes volumetric data through hierarchical resolution stages with skip connections to preserve fine-grained spatial features. Unlike prior transformer-based models or heavily modular CNNs, our design favors plain convolution blocks to minimize computational burden while preserving capacity for large-scale context modeling. Following insights from prior work~\cite{ding2022scaling}, we adopt a $21 \times 21 \times 21$ depthwise convolution (DWC-21) as the kernel backbone, which we empirically identify as the best trade-off between expressiveness and efficiency in 3D. Each encoder block consists of batch normalization, followed by the depthwise convolution and GELU activation. The feature propagation from layer $\ell-1$ to layer $\ell$ and then to $\ell+1$ is defined as:
\begin{align}
    \hat{z}_\ell &= \mathrm{GELU}\big(\mathrm{DWC}_{21}(\mathrm{BN}(z_{\ell-1}))\big), \\
    \hat{z}_{\ell+1} &= \mathrm{GELU}\big(\mathrm{DWC}_{21}(\mathrm{BN}(\hat{z}_\ell))\big),
\end{align}
where $z_{\ell-1}$ is the input from the previous layer, $\hat{z}_\ell$ and $\hat{z}_{\ell+1}$ are intermediate representations, $\mathrm{BN}$ denotes batch normalization, and $\mathrm{DWC}_{21}$ represents depthwise convolution with a $21^3$ kernel. This architectural choice allows the network to efficiently encode both local and global context, while enabling seamless integration of our re-parameterized learning framework.

\section{Experimental Setup}
\textbf{Datasets and Implementation Details.} We evaluate Rep3D on five publicly available volumetric segmentation datasets, covering a wide range of anatomical structures across different spatial scales, from large organs (e.g., liver, stomach) to smaller and more challenging targets (e.g., tumors, vessels). We report results using the Dice Similarity Coefficient (DSC) as the primary evaluation metric, quantifying spatial overlap between predicted segmentations and ground truth labels. Additional details, including dataset resolution normalization, voxel spacing, and pre-processing pipelines and experimental details are provided in the appendix.

\begin{table*}[t!]
    \centering
    \caption{Comparison of SOTA approaches on the three different testing datasets. (*: $p< 0.01$, with Paired Wilcoxon signed-rank test to all baseline networks)}
    \label{tab:sota_compare}
    \begin{adjustbox}{width=\textwidth}
    \begin{tabular}{lccccccccccccc}
        \toprule
         & & & \multicolumn{4}{c}{KiTS} & \multicolumn{3}{c}{MSD Pancreas} & \multicolumn{3}{c}{MSD Hepatic} \\
         \cmidrule(lr){4-7} \cmidrule(lr){8-10} \cmidrule(lr){11-13}
        Methods & \#Params & FLOPs & Kidney & Tumor & Cyst & Mean & Pancreas & Tumor & Mean & Hepatic & Tumor & Mean \\
         \midrule
         3D U-Net \cite{cciccek20163d} & 4.81M & 135.9G & 0.918 & 0.657 & 0.361 & 0.645 & 0.711 & 0.584 & 0.648 & 0.569 & 0.609 & 0.589\\
         SegResNet \cite{myronenko20183d} & 1.18M & 15.6G & 0.935 & 0.713 & 0.401 & 0.683 & 0.740 & 0.613 & 0.677 & 0.620 & 0.656 & 0.638\\
         RAP-Net \cite{lee2021rap} & 38.2M & 101.2G & 0.931 & 0.710 & 0.427 & 0.689 & 0.742 & 0.621 & 0.682 & 0.610 & 0.643 & 0.627\\
         nn-UNet \cite{isensee2021nnu} & 31.2M & 743.3G & 0.943 & 0.732 & 0.443 & 0.706 & 0.775 & 0.630 & 0.703 & 0.623 & 0.695 & 0.660\\
         \midrule
         TransBTS \cite{wang2021transbts} & 31.6M & 110.3G & 0.932 & 0.691 & 0.384 & 0.669 & 0.749 & 0.610 & 0.679 & 0.589 & 0.636 & 0.613\\ 
         UNETR \cite{hatamizadeh2022unetr} & 92.8M & 82.5G & 0.921 & 0.669 & 0.354 & 0.648 & 0.735 & 0.598 & 0.667 & 0.567 & 0.612 & 0.590\\
         nnFormer \cite{zhou2021nnformer} & 149.3M & 213.0G & 0.930 & 0.687 & 0.376 & 0.664 & 0.769 & 0.603 & 0.686 & 0.591 & 0.635 & 0.613\\ 
         SwinUNETR \cite{hatamizadeh2022swin} & 62.2M & 328.1G & 0.939 & 0.702 & 0.400 & 0.680 & 0.785 & 0.632 & 0.708 & 0.622 & 0.647 & 0.635\\
         3D UX-Net (k=7) \cite{lee20223d} & 53.0M & 639.4G & 0.942 & 0.724 & 0.425 & 0.697 & 0.737 & 0.614 & 0.676 & 0.625 & 0.678 & 0.652 \\
         UNesT-B \cite{yu2023unest} & 87.2M & 258.4G & 0.943 & 0.746 & 0.451 & 0.710  & 0.778 & 0.601 & 0.690 & 0.611 & 0.645 & 0.640 \\
         \midrule
         Rep3D (Fixed Prior) & 65.8M & 757.4G & 0.950 & 0.757 & 0.473 & 0.727 & 0.789 & 0.640 & 0.715 & 0.635 & 0.681 & 0.658\\
         Rep3D & 66.0M & 757.6G & \textbf{0.955} & \textbf{0.763} & \textbf{0.490} & \textbf{0.736*} & \textbf{0.793} & \textbf{0.653} & \textbf{0.723*} & \textbf{0.650} & \textbf{0.697} & \textbf{0.674*} \\
         \bottomrule
    \end{tabular}
    \end{adjustbox}
\end{table*}

\begin{table*}[t!]
    \caption{Evaluations on the AMOS testing split in different scenarios (*: $p< 0.01$).}
    \label{tab:amos_eval}
    \begin{adjustbox}{width=\textwidth}
    \begin{tabular}{lccccccccccccccc|c}
        \toprule
        \multicolumn{17}{c}{\textbf{AMOS CT (Train From Scratch Scenario)}} \\ 
        \midrule
        Methods & Spleen & R.Kid & L.Kid & Gall. & Eso. & Liver & Stom. & Aorta & IVC & Panc. & RAG & LAG & Duo. & Blad. & Pros. & Avg\\ 
        \midrule
        nn-UNet (350 Ep) & 0.951 & 0.919 & 0.930 & 0.845 & 0.797 & 0.975 & 0.863 & 0.941 & 0.898 & 0.813 & 0.730 & 0.677 & 0.772 & 0.797 & 0.815 & 0.850 \\
        nn-UNet (1000 Ep) & 0.967 & 0.958 & 0.945 & 0.890 & 0.818 & 0.979 & 0.914 & 0.953 & 0.920 & 0.824 & 0.799 & 0.743 & 0.823 & 0.900 & 0.867 & 0.887 \\
        \midrule
        TransBTS & 0.930 & 0.921 & 0.909 & 0.798 & 0.722 & 0.966 & 0.801 & 0.900 & 0.820 & 0.702 & 0.641 & 0.550 & 0.684 & 0.730 & 0.679 & 0.783 \\
        UNETR & 0.925 & 0.923 & 0.903 & 0.777 & 0.701 & 0.964 & 0.759 & 0.887 & 0.821 & 0.687 & 0.688 & 0.543 & 0.629 & 0.710 & 0.707 & 0.740 \\
        nnFormer & 0.932 & 0.928 & 0.914 & 0.831 & 0.743 & 0.968 & 0.820 & 0.905 & 0.838 & 0.725 & 0.678 & 0.578 & 0.677 & 0.737 & 0.596 & 0.785 \\
        SwinUNETR & 0.956 & 0.957 & 0.949 & 0.891 & 0.820 & 0.978 & 0.880 & 0.939 & 0.894 & 0.818 & 0.800 & 0.730 & 0.803 & 0.849 & 0.819 & 0.871 \\ 
        3D UX-Net (k=7) & 0.966 & 0.959 & 0.951 & 0.903 & 0.833 & 0.980 & 0.910 & 0.950 & 0.913 & 0.830 & 0.805 & 0.756 & 0.846 & 0.897 & 0.863 & 0.890 \\ 
        3D UX-Net (k=21) & 0.963 & 0.959 & 0.953 & 0.921 & 0.848 & 0.981 & 0.903 & 0.953 & 0.910 & 0.828 & 0.815 & 0.754 & 0.824 & 0.900 & 0.878 & 0.891 \\ 
        UNesT-B & 0.966 & 0.961 & 0.956 & 0.903 & 0.840 & 0.980 & 0.914 & 0.947 & 0.912 & 0.838 & 0.803 & 0.758 & 0.846 & 0.895 & 0.854 & 0.891\\
        RepOptimizer & 0.968 & 0.964 & 0.953 & 0.903 & 0.857 & 0.981 & 0.915 & 0.950 & 0.915 & 0.826 & 0.802 & 0.756 & 0.813 & 0.906 & 0.867 & 0.892 \\
        \midrule
        Rep3D (Fixed) & 0.972 & 0.963 & 0.964 & 0.911 & 0.861 & 0.982 & 0.921 & 0.956 & 0.924 & 0.837 & 0.818 & 0.777 & 0.831 & 0.916 & 0.879 & 0.902 \\ 
        Rep3D (LRBM) & \textbf{0.978} & \textbf{0.970} & \textbf{0.964} & \textbf{0.928} & \textbf{0.871} & \textbf{0.984} & \textbf{0.927} & \textbf{0.960} & \textbf{0.930} & \textbf{0.851} & \textbf{0.828} & \textbf{0.784} & \textbf{0.850} & \textbf{0.920} & \textbf{0.881} & \textbf{0.910*} \\ 
        \midrule
        \multicolumn{17}{c}{\textbf{AMOS MRI (Train From Scratch Scenario)}} \\ 
        \midrule
        Methods & Spleen & R.Kid & L.Kid & Gall. & Eso. & Liver & Stom. & Aorta & IVC & Panc. & RAG & LAG & Duo. & Blad. & Pros. & Avg\\ 
        \midrule
        nn-UNet (350 Ep) & 0.967 & 0.855 & 0.958 & 0.663 & 0.736 & 0.973 & 0.888 & 0.956 & 0.907 & 0.793 & 0.533 & 0.572 & 0.668 & - & - & 0.805 \\
        nn-UNet (1000 Ep) & 0.973 & 0.940 & 0.965 & \textbf{0.681} & 0.810 & 0.980 & \textbf{0.893} & 0.967 & \textbf{0.917} & 0.834 & 0.667 & 0.689 & 0.701 & - & - & 0.847 \\
        \midrule
        TransBTS & 0.956 & 0.957 & 0.955 & 0.619 & 0.770 & 0.974 & 0.867 & 0.958 & 0.852 & 0.836 & 0.591 & 0.630 & 0.648 & - & - & 0.816 \\
        UNETR & 0.942 & 0.956 & 0.930 & 0.552 & 0.741 & 0.967 & 0.836 & 0.947 & 0.829 & 0.815 & 0.564 & 0.621 & 0.624 & - & - & 0.794 \\
        nnFormer & 0.949 & 0.952 & 0.950 & 0.601 & 0.758 & 0.972 & 0.859 & 0.960 & 0.843 & 0.832 & 0.569 & 0.618 & 0.637 & - & - & 0.808 \\
        SwinUNETR & 0.972 & 0.961 & 0.961 & 0.649 & 0.814 & 0.978 & 0.889 & 0.961 & 0.862 & 0.854 & 0.659 & 0.649 & 0.664 & - & - & 0.836 \\ 
        3D UX-Net (k=7) & 0.971 & 0.965 & 0.966 & 0.603 & 0.828 & 0.978 & 0.869 & 0.962 & 0.878 & 0.837 & 0.696 & 0.689 & 0.696 & - & - & 0.841 \\
        3D UX-Net (k=21) & 0.968 & 0.962 & 0.967 & 0.610 & 0.830 & 0.977 & 0.858 & 0.954 & 0.880 & 0.829 & 0.701 & 0.697 & 0.700 & - & - & 0.840\\
        UNesT-B & 0.971 & 0.965 & 0.967 & 0.615 & 0.831 & 0.980 & 0.865 & 0.949 & 0.883 & 0.845 & 0.691 & 0.700 & 0.697 & - & - & 0.843\\
        RepOptimizer & 0.970 & 0.967 & 0.971 & 0.635 & 0.823 & 0.978 & 0.875 & 0.963 & 0.882 & 0.850 & 0.689 & 0.691 & 0.711 & - & - & 0.847 \\
        \midrule
        Rep3D (Fixed) & 0.972 & 0.965 & 0.970 & 0.644 & 0.838 & 0.980 & 0.883 & 0.965 & 0.893 & 0.861 & 0.714 & 0.701 & 0.725 & - & - & 0.855 \\ 
        Rep3D (LRBM) & \textbf{0.975} & \textbf{0.969} & \textbf{0.975} & 0.657 & \textbf{0.845} & \textbf{0.984} & 0.891 & \textbf{0.970} & 0.901 & \textbf{0.879} & \textbf{0.718} & \textbf{0.721} & \textbf{0.750} & - & - & \textbf{0.864*} \\ 
        \bottomrule
    \end{tabular}
    \end{adjustbox}
\end{table*}

\section{Results}
\subsection{Evaluation on Tissue \& Tumor Segmentation}
To assess the generalization and scalability of Rep3D across diverse anatomical structures and clinical targets, we evaluate performance on three representative volumetric segmentation tasks using the KiTS, MSD Pancreas, and MSD Hepatic Vessel datasets. As shown in Table \ref{tab:sota_compare}, Rep3D achieves state-of-the-art performance across all settings, consistently outperforming both convolution- and transformer-based baselines. On the KiTS dataset, which includes kidney, tumor, and cyst segmentation, Rep3D achieves the highest average Dice score of 0.736, with strong individual scores of 0.955 (kidney), 0.763 (tumor), and 0.490 (cyst). Notably, Rep3D improves tumor segmentation performance by 2.28\% Dice over UNesT-B and 5.39\% Dice over 3D UX-Net, demonstrating its ability to adapt to complex local variations in pathological regions. On the MSD Pancreas task, which is particularly challenging due to the pancreas's low contrast and irregular boundaries, Rep3D sets a new benchmark with an average Dice score of 0.723, outperforming SwinUNETR (0.708), nnUNet (0.703), and UNesT-B (0.690). Tumor segmentation also benefits from our re-parameterization design, improving by 3.32\% Dice compared to 3D UX-Net and 2.03\% Dice compared to the fixed-prior variant. On the MSD Hepatic Vessel dataset, Rep3D continues to lead with a mean Dice of 0.674, outperforming the previous best model (UNesT-B, 0.640) and demonstrating superior vessel and tumor localization. The results also highlight the effectiveness of Rep3D’s spatially adaptive learning dynamics, especially in sparse and small-structure segmentation where traditional large-kernel convolutions or global self-attention tend to underperform.
\begin{figure*}[t!]
\centering
\includegraphics[width=\textwidth]{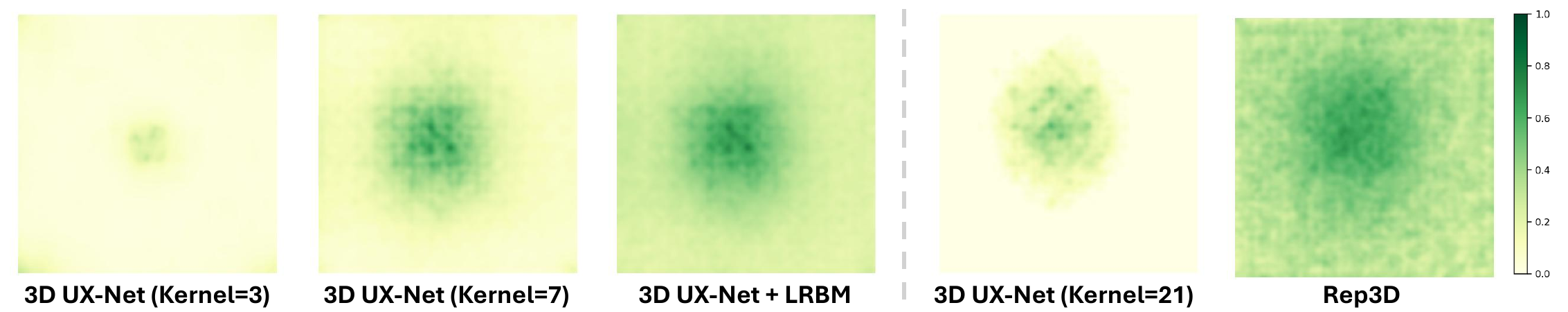}
\caption{As kernel size increases, depthwise convolutions in 3D UX-Net exhibit increasingly diffuse ERFs, gradually expanding the gradient dynamics from local to broader spatial regions. Incorporating LRBM further enhances weighting toward global areas by modulating the spatial contribution of distant elements. In contrast, Rep3D produces a well-distributed ERF that preserves strong central activation while extending contextual influence across the full kernel.} \label{fig:erf_figure}
\end{figure*}
\begin{table*}[t!]
    \caption{Ablation Studies on bias generator's convolutional layers and LRBM 3D adaptability.}
    \label{tab:ablation}
    \begin{adjustbox}{width=\textwidth}
    \begin{tabular}{lccccccccccccccc|c}
        \toprule
        Methods & Spleen & R.Kid & L.Kid & Gall. & Eso. & Liver & Stom. & Aorta & IVC & Panc. & RAG & LAG & Duo. & Blad. & Pros. & Avg\\ 
        \midrule
        Kernel=$1\times1\times1$ & 0.972 & 0.968 & \textbf{0.965} & 0.926 & 0.863 & 0.984 & 0.917 & 0.956 & 0.922 & 0.851 & 0.816 & 0.779 & \textbf{0.863} & 0.912 & \textbf{0.894} & 0.905\\
        Kernel=$3\times3\times3$ & 0.970 & 0.966 & 0.960 & \textbf{0.930} & 0.863 & 0.984 & \textbf{0.935} & 0.958 & 0.924 & \textbf{0.859} & 0.827 & 0.758 & 0.862 & 0.908 & 0.892 & 0.906 \\
        Kernel=$5\times5\times5$ & 0.974 & 0.967 & 0.964 & 0.925 & 0.833 & 0.984 & 0.924 & 0.956 & 0.910 & 0.850 & \textbf{0.829} & \textbf{0.786} & 0.843 & \textbf{0.921} & 0.884 & 0.903 \\
        Kernel=$7\times7\times7$ & \textbf{0.978} & \textbf{0.970} & 0.964 & 0.928 & \textbf{0.871} & \textbf{0.984} & 0.927 & \textbf{0.960} & \textbf{0.930} & 0.851 & 0.828 & 0.784 & 0.850 & 0.920 & 0.881 & \textbf{0.910} \\
        \midrule
        3D UX-Net (k=7) & 0.966 & 0.959 & 0.951 & 0.903 & 0.833 & 0.980 & 0.910 & 0.950 & 0.913 & 0.830 & 0.805 & 0.756 & 0.846 & 0.897 & 0.863 & 0.890 \\ 
        3D UX-Net + LRBM & 0.968 & 0.963 & 0.952 & 0.911 & 0.841 & 0.981 & 0.915 & 0.959 & 0.920 & 0.835 & 0.811 & 0.770 & 0.851 & 0.901 & 0.872 & 0.897 \\
        \bottomrule
    \end{tabular}
    \end{adjustbox}
\end{table*}

\subsection{Evaluation on Multi-Organ Segmentation}
Beyond the ability to segment anatomical structures across scales, we further evaluate Rep3D on the AMOS benchmark under the ``train-from-scratch'' setting for both CT and MRI modalities. On AMOS-CT, Rep3D achieves the best performance across all 15 evaluated anatomical structures, surpassing strong baselines including SwinUNETR, UNesT, and 3D UX-Net. Notably, Rep3D outperforms UNesT-B by 2.13\% and RepOptimizer by 2.02\% of average Dice score, while operating with fewer parameters than UNesT. On AMOS-MRI, a more challenging modality due to the variable range of contrast intensity and anatomical ambiguity, Rep3D maintains its superior performance, achieving an average Dice of 0.864, again outperforming all competing approaches. Compared to the best-performing transformer baseline (UNesT-B, 0.854) and convolutional baseline (3D UX-Net (k=21), 0.840), Rep3D delivers consistent improvements across nearly all organ classes, particularly in difficult regions such as the pancreas, gallbladder, and adrenal glands. These gains underscore the effectiveness of our spatially adaptive re-parameterization strategy in enhancing convergence and feature expressivity without increasing model complexity.

\subsection{Ablation Studies}
\textbf{Isolation of Spatial Modulation (LRBM vs. Vanilla).} To strictly isolate the contribution of our proposed Low-Rank Receptive Bias Modeling (LRBM), we compare Rep3D against a ``Vanilla Rep'' baseline (Appendix Table 7). This baseline utilizes the identical plain encoder architecture (a single $21\times21\times21$ branch per block) but is trained using standard stochastic gradient descent without spatial modulation. The results are decisive: the Vanilla Rep baseline exhibits significant optimization instability, characterized by high variance in training loss and consistently lower Dice scores. This empirical failure confirms that the ``plain'' large-kernel architecture is inherently unstable on volumetric data; our learnable spatial prior is not merely an enhancement, but a necessary condition for convergence in this regime.

\textbf{Optimal Capacity for Bias Modeling (Generator Depth).} We investigate the representational capacity required to model the spatial gradient prior effectively. By varying the depth of the generator network $f_\theta$ (from 1 to 3 layers), we observe a clear trade-off between expressivity and optimizability. The 1-layer generator proves insufficient, likely because a linear mapping cannot capture the complex, non-isotropic diffusion patterns of the effective receptive field. Conversely, the 3-layer variant introduces optimization difficulties, leading to a slight performance degradation ($0.910 \to 0.899$ Dice). Our results identify the 2-layer configuration as the optimal design, providing sufficient non-linearity to model nuanced spatial priors while maintaining a lightweight footprint that does not impede gradient flow.

\textbf{Local vs. Global Context in Bias Generation (Generator Kernel Size).} While the Rep3D backbone utilizes a massive $21\times21\times21$ kernel to capture global semantic context, the spatial bias generator may operate on a different scale. We analyzed the generator's kernel size ($K_G \in \{1^3, 3^3, 5^3, 7^3\}$) to determine the optimal receptive field for determining gradient weights. As shown in Table \ref{tab:ablation}, the impact is organ-dependent. Small generator kernels ($1^3$) favor boundary-sensitive organs (e.g., bladder), suggesting that fine-grained local statistics are crucial for edge delineation. However, the $7\times7\times7$ generator achieves the highest overall Dice (0.910), indicating that a larger receptive field is the best for estimating the spatial prior. This suggests that the decision to up-weight or down-weight a central pixel should be informed by its immediate neighborhood, even if the backbone filter itself spans a much larger volume.

\textbf{Universality of LRBM (Integration with 3D UX-Net).} Finally, to demonstrate that our findings are not specific to the Rep3D architecture, we integrate the LRBM module into a standard 3D UX-Net (which uses fixed $7\times7\times7$ kernels). As reported in Table \ref{tab:ablation}, this integration improves the average Dice score from 0.890 to 0.897. The gains are particularly pronounced in anatomically challenging regions with irregular shapes, such as the left adrenal gland ($0.756 \to 0.770$) and the duodenum ($0.846 \to 0.851$). These improvements confirm that spatially adaptive learning rates provide a robust inductive bias that aids convergence in hard-to-segment regions, functioning as a plug-and-play enhancement for existing volumetric segmentation backbones.

\section{Discussion \& Limitations}
In this work, we introduced Rep3D, a re-parameterization framework that explicitly models spatial convergence dynamics in large kernel 3D convolutions. By linking effective receptive field (ERF) behavior with first-order optimization theory, we demonstrated that large convolution kernels naturally exhibit non-uniform learning dynamics, where central elements converge faster than peripheral ones. To address this, Rep3D integrates a learnable spatial prior via low-rank modulation, allowing the optimizer to differentially emphasize kernel regions with the distinctive characteristics of ERF during training. Our experiments across five diverse 3D segmentation benchmarks, confirm that Rep3D consistently improves performance over both transformer-based and convolution-based SOTA approaches, while maintaining a plain and efficient encoder design. The success of Rep3D reinforces several broader insights. First, spatially adaptive optimization is a promising direction for bridging inductive biases in CNNs with the dynamic learning capacity of attention-based models. Second, incorporating explicit ERF modeling into kernel design enables more efficient parameter usage, particularly in data-limited medical imaging scenarios. Moreover, our framework enhance network interpretability: the modulation masks can be visualized and aligned with ERF pattern as demonstrated in Figure \ref{fig:erf_figure}, offering insights into how spatial understanding guides the learning of convolution kernels.

While Rep3D demonstrates strong empirical performance across diverse 3D medical segmentation tasks, several limitations remain. First, although our learnable modulation mechanism introduces minimal architectural overhead, the training cost associated with large 3D kernels (e.g., $21\times21\times21$) remains nontrivial, particularly in memory-constrained GPU environments. Unlike 2D convolution kernels (i.e. MegEngine packages for 2D depthwise kernels), limited packages and approaches have been proposed to optimize the large kernel mechanism in 3D. This limits the batch size and input resolution during training, which can affect convergence and generalization. Future work could explore progressive training strategies, multi-resolution optimization, or low-resolution proxy supervision to alleviate this constraint while maintaining segmentation fidelity. Second, while our distance decay prior effectively guides spatial re-parameterization, its performance is inherently tied to the input volume resolution. In our experiments, we downsample 3D volumes to specific resolution (e.g., $1.5 \times 1.5 \times 2.0$ mm) to balance computation and efficiency. However, we observe saturation effects when training at higher resolutions, where further improvements in image quality do not yield proportional gains in segmentation accuracy. This may be due to the spatial prior losing precision at finer scales. Adapting fine-grained spatial learnable prior could be another potential direction for future work.

\section{Conclusion}
In this paper, we introduced Rep3D, a receptive-biased re-parameterization framework for large kernel 3D convolutions. By modeling effective receptive field (ERF) behavior as a learnable spatial prior, Rep3D enables adaptive element-wise learning dynamics during training, bridging the gap between convolutional inductive bias and optimization-aware design. Implemented via a lightweight modulation network, our approach avoids complex multi-branch architectures while improving training efficiency and segmentation accuracy. Extensive experiments across five volumetric medical imaging benchmarks demonstrate consistent improvements over SOTA transformer and CNN approaches, establishing Rep3D as a scalable and effective solution for 3D medical image analysis.



\bibliography{icml2026}
\bibliographystyle{icml2026}

\newpage
\appendix
\onecolumn
\section{Rep3D Motivations from Experiments: Naively Scaling 3D Convolution Kernels}
Recent advances in medical image segmentation have highlighted the importance of large receptive fields for capturing long-range spatial dependencies in volumetric data. Motivated by this, there has been a growing trend toward enlarging convolutional kernel sizes in 3D CNN architectures, such as 3D UX-Net~\cite{lee20223d} and RepUX-Net~\cite{lee2023scaling}, which attempt to mimic the global context modeling capabilities of transformers while retaining the inductive biases of CNNs.

However, the straightforward enlargement of kernel size introduces several practical and theoretical challenges:
\begin{itemize}
    \item \textbf{Optimization Instability:} Large convolutional kernels suffer from slow or uneven convergence, particularly in the outer kernel regions, which are rarely activated in early training. This leads to ineffective utilization of capacity and suboptimal learning behavior.
    \item \textbf{Degrading Performance:} Empirically, simply increasing the kernel size does not guarantee improved performance. Beyond a certain scale, performance tends to saturate or even degrade.
    \item \textbf{Inefficient Parameter Usage:} Naïve kernel scaling introduces a quadratic (especially in 3D) growth in the number of parameters, making training inefficient and difficult to regularize.
\end{itemize}

To illustrate these effects empirically, we conduct a systematic ablation study on the AMOS dataset using the 3D UX-Net encoder as a backbone. 3D UX-Net is an ideal starting point because it already leverages larger kernels ($7 \times 7 \times 7$) in its baseline. We vary the convolutional kernel size from $3 \times 3 \times 3$ to $21 \times 21 \times 21$, keeping all other architectural and training settings fixed. The results, shown in Table 4 , confirm our hypothesis.

\begin{table}[H]
\centering
\caption{Impact of kernel size on segmentation performance using 3D UX-Net encoder on the AMOS dataset.}
\label{tab:kernel_scaling_ablation}
\begin{tabular}{c|c}
\toprule
\textbf{Kernel Size} & \textbf{Avg. Dice Score} \\
\midrule
$3 \times 3 \times 3$  & 0.881 \\
$5 \times 5 \times 5$  & 0.885 \\
$7 \times 7 \times 7$  & 0.890 \\
$9 \times 9 \times 9$  & 0.891 \\
$11 \times 11 \times 11$ & 0.893 \\
$13 \times 13 \times 13$ & 0.894 \\
$15 \times 15 \times 15$ & \textbf{0.895} \\
$17 \times 17 \times 17$ & 0.893 \\
$19 \times 19 \times 19$ & 0.893 \\
$21 \times 21 \times 21$ & 0.891 \\
\bottomrule
\end{tabular}
\end{table}

As seen in Table 4, performance initially improves with increasing kernel size, peaking at $15 \times 15 \times 15$. However, further enlargements yield diminishing or even negative returns, despite increased computational cost. These findings reveal a key limitation of naïve kernel enlargement: Although it increases theoretical receptive field, it fails to translate into meaningful gains due to the lack of spatially adaptive optimization.

\subsection{Theoretical extension from SGD to Adam/AdamW}
We provide a rigorous derivation of how the Constant-Scale Linear Addition (CSLA) block induces a spatially varying learning rate, specifically under adaptive optimization algorithms like Adam (Kingma \& Ba, 2014).

\textbf{1. The Normalization Cancellation Effect}
Consider the update rule for a single parameter $w$ in Adam. The update $\Delta w_t$ at step $t$ is given by:
\begin{equation}
    \Delta w_t = -\eta \cdot \frac{m_t}{\sqrt{v_t} + \epsilon},
\end{equation}
where $m_t$ is the first moment (momentum) and $v_t$ is the second moment (variance) of the gradients. 
In a re-parameterized branch scaled by $\alpha$ (i.e., $w_{\text{branch}} = \alpha w$), the gradient is scaled by $\alpha$: $g \leftarrow \alpha g$.
Consequently, the moments scale as $m_t \leftarrow \alpha m_t$ and $v_t \leftarrow \alpha^2 v_t$. Substituting these into the update rule:
\begin{equation}
    \Delta w_{\text{branch}} = -\eta \cdot \frac{\alpha m_t}{\sqrt{\alpha^2 v_t} + \epsilon} \approx -\eta \cdot \frac{m_t}{\sqrt{v_t}}.
\end{equation}
As noted, the scaling factor $\alpha$ cancels out. This implies that the magnitude of the update step for any single branch is largely independent of its scaling factor $\alpha$. It is effectively governed only by the global learning rate $\eta$.

\textbf{2. Structural Superposition and Effective Step Size}
However, such derivation lies in analyzing branches in isolation. The effective kernel $W'$ used during inference is the structural superposition of the branches:
\begin{equation}
    W' = \alpha_L W_L + \alpha_S \mathcal{P}(W_S),
\end{equation}
where $\mathcal{P}(\cdot)$ represents the zero-padding operator that aligns the small kernel to the spatial dimensions of the large kernel.

The effective update $\Delta W'$ is the linear combination of the branch updates:
\begin{equation}
    \Delta W'(x) = \alpha_L \Delta W_L(x) + \alpha_S \Delta W_S(x).
\end{equation}
We can further analyze the magnitude of this effective update across spatial positions $x$:

\textbf{Case A: Peripheral Region ($x \in \text{Periphery}$)}
These spatial positions exist only in the support of $W_L$. The effective update is:
\begin{equation}
    \Delta W'_{\text{peri}} = \alpha_L \Delta W_L \approx \alpha_L \cdot \eta \cdot \vec{u}_L,
\end{equation}
where $\vec{u}_L$ is the normalized update direction. The step size magnitude is proportional to one unit of update.

\textbf{Case B: Central Region ($x \in \text{Center}$)}
These spatial positions exist in the support of \textit{both} $W_L$ and $W_S$. The effective update is the superposition of both branches:
\begin{equation}
    \Delta W'_{\text{center}} = \alpha_L \Delta W_L + \alpha_S \Delta W_S \approx \alpha_L (\eta \vec{u}_L) + \alpha_S (\eta \vec{u}_S).
\end{equation}
Assuming the gradient directions $\vec{u}_L$ and $\vec{u}_S$ are not perfectly orthogonal (they are optimizing the same loss on the same features), these vectors constructively interfere.
Crucially, even if Adam normalizes $\Delta W_L$ and $\Delta W_S$ to have similar magnitudes, the center receives the sum of two update vectors, while the periphery receives only one.

\textbf{3. The Induced Spatial Prior}
This derivation proves that structurally re-parameterized blocks inherently induce a spatial learning rate field $\lambda(x)$:
\begin{equation}
    \lVert \Delta W'(x) \rVert \propto 
    \begin{cases} 
    C_1 \cdot \eta & \text{if } x \in \text{Periphery} \\
    (C_1 + C_2) \cdot \eta & \text{if } x \in \text{Center}
    \end{cases}
\end{equation}
where $C_1, C_2 > 0$ are constants derived from the structural combination. This confirms that optimization is faster at the center and slower at the periphery, aligning with the "Gaussian" diffusion pattern of Effective Receptive Fields (ERFs), demonstrating similar outcome with the SGD optimizer.

\subsection{Data Preprocessing \& Training Details}
\begin{table*}[h]
    \centering
    \caption{Hyperparameters for direction training scenario on four public datasets}
    \begin{adjustbox}{width=0.7\textwidth}
    \fontsize{9pt}{10pt}\selectfont
    \begin{tabular}{*{1}{l}|*{1}{c}}
        \toprule
        \textbf{Hyperparameters} & \textbf{Direct Training}  \\
        \midrule
        Encoder Stage & 4 \\
        Layer-wise Channel & $48, 96, 192, 384$ \\
        Hidden Dimensions & $768$ \\
        Patch Size & $96\times96\times96$ \\
        No. of Sub-volumes Cropped & 2 \\
        \midrule
        Training Steps & 60000\\
        Batch Size & 1\\
        AdamW $\epsilon$ & $1e-8$\\
        AdamW $\beta$ & ($0.9, 0.999$)\\
        Peak Learning Rate & $1e-4$\\
        Learning Rate Scheduler & ReduceLROnPlateau \\
        Factor \& Patience & 0.9, 10\\
        \midrule
        Dropout & X\\
        Weight Decay & 0.08 \\
        \midrule
        Data Augmentation & Intensity Shift, Rotation, Scaling\\
        Cropped Foreground & \checkmark\\
        Intensity Offset & 0.1 \\
        Rotation Degree & $-30^{\circ}$ to $+30^{\circ}$ \\
        Scaling Factor & x: 0.1, y: 0.1, z: 0.1 \\
         \bottomrule
    \end{tabular}
    \end{adjustbox}
    \label{baselines_compare}
\end{table*}

We apply hierarchical steps for data preprocessing: 1) intensity clipping is applied to further enhance the contrast of soft tissue (AMOS CT, KiTS, MSD Pancreas:\{min:-175, max:250\}; MSD Hepatic Vessel:\{min:0, max:230\}); AMOS MRI:\{min:0, max:1000\}. 2) Intensity normalization is performed after clipping for each volume and use min-max normalization: $(X-X_1)/(X_{99}-X_1)$ to normalize the intensity value between 0 and 1, where $X_p$ denote as the $p_{th}$ percentile of intensity in $X$. We then perform downsampling to certain voxel spacing (i.e. AMOS CT, MSD hepatic vessels, MSD Pancreas and KiTS: $1.5\times1.5\times2.0$, AMOS MRI: $1.0\times1.0\times1.0$) randomly crop sub-volumes with size $96\times96\times96$ at the foreground and perform data augmentations, including rotations, intensity shifting, and scaling (scaling factor: 0.1). All training processes with Rep3D are optimized with either Stochastic Gradient Descent (SGD) or AdamW optimizer. We trained all models for 60000 steps using a learning rate of $0.0001$ on an NVIDIA A100 GPU across all datasets. One epoch takes approximately about 9 minutes for KiTS, 5 minutes for MSD Pancreas, 12 minutes for MSD hepatic vessels, 7 minutes for AMOS CT and 1 minute for AMOS MRI, respectively. 

All experiments are conducted under a direct supervised learning setting. For the KiTS and MSD datasets, we employ a 5-fold cross-validation strategy using an 80\%/10\%/10\% split for training, validation, and testing, respectively. For the AMOS dataset, we use a fixed single split with the same partitioning ratio. Details on training procedures and preprocessing protocols are provided in the supplementary material. Our proposed re-parameterization approach Rep3D, is benchmarked against both convolutional and transformer-based state-of-the-art (SOTA) methods for 3D medical image segmentation. For nnUNet \cite{isensee2021nnu}, we evaluate performance across two different training schedules to account for fairness, since Rep3D is trained with 60,000 iterations (approximately equivalent to 350 epochs). Therefore, we have provided performance with partial scheduled (350 epochs) and full scheduled (1000 epochs) to demonstrate our model generalizability. 

\subsection{Datasets Details}
We have leverage four challenging public datasets across different scales: 1) AMOS22 (MICCAI 2022 Abdominal Multi-organ Segmentation Challenge) \cite{ji2022amos}: Comprises 200 multi-contrast abdominal CT scans with 15 organ-level anatomical labels and 33 MRI scans with 13 organ-level anatomical labels for comprehensive abdominal segmentation, 2) KiTS21 (MICCAI 2021 Kidney Tumor Segmentation Challenge) \cite{heller2019kits19}: Includes 210 contrast-enhanced abdominal CT scans from the University of Minnesota Medical Center (2010–2018), with manual annotations for kidney, tumor, and cyst, 3) MSD Pancreas (Medical Segmentation Decathlon) \cite{antonelli2022medical}: Contains 282 abdominal contrast-enhanced CT scans annotated for both pancreas and pancreatic tumor segmentation, and 4) MSD Hepatic Vessel (Medical Segmentation Decathlon) \cite{antonelli2022medical}: Contains 303 abdominal CT scans annotated for hepatic vessel and associated tumor segmentation.

\begin{table*}[h]
    \centering
    \caption{Complete overview of Four public datasets}
    \begin{adjustbox}{width=\textwidth}
    \begin{tabular}{*{1}{l}|*{5}{c}}
        \toprule
        Challenge & AMOS CT & AMOS MR & MSD Pancreas & MSD Hepatic Vessels & KiTS \\
        \midrule
        Imaging Modality & Multi-Contrast CT & Multi-Contrast MRI & \multicolumn{2}{c}{Venous CT} &  Arterial CT \\
        Anatomical Region & \multicolumn{2}{c}{Abdomen} & Pancreas & Liver & Kidney\\
        Sample Size & 200 & 33 & 282 & 303 & 300 \\
        \midrule
        \multirow{4}{*}{Anatomical Label} & \multicolumn{2}{c}{Spleen, Left \& Right Kidney, Gall Bladder,} &  \multirow{4}{*}{Pancreas, Tumor} & \multirow{4}{*}{Hepatic Vessels, Tumor} &  \multirow{4}{*}{Kidney, Tumor} \\
        & \multicolumn{2}{c}{Esophagus, Liver, Stomach, Aorta, Inferior Vena Cava (IVC)} & &  \\
        & \multicolumn{2}{c}{Pancreas, Left \& Right Adrenal Gland (AG), Duodenum} & & \\
        & \multicolumn{2}{c}{Bladder (CT only), Prostate/Uterus (CT only)} & & \\
        \midrule
        \multirow{2}{*}{Data Splits}  & \multicolumn{2}{c}{1-Fold (Internal)} & \multicolumn{3}{c}{5-Fold Cross-Validation}\\
        & Train: 160 / Validation: 20 / Test: 20 & Train: 22 / Validation: 4/ Test: 7 & Train: 225 / Validation: 27 / Testing: 30 & Training: 242, Validation: 30 / Testing: 31 & Training: 240, Validation: 30 / Testing: 30\\
        \midrule
        5-Fold Ensembling & N/A & N/A & X & \checkmark & X \\
        \bottomrule
    \end{tabular}
    \end{adjustbox}
    \label{baselines_compare}
\end{table*}

\subsection{Network Architecture}
We adopt a 3D encoder-decoder architecture from both 3D UX-Net \cite{lee20223d} and SwinUNETR \cite{hatamizadeh2022swin} as the backbone of Rep3D. Instead of using encoder block with feed forward layer, we simply using a plain convolutional design with depthwise separable convolutions in parallel with LRBM. The encoder consists of 4 hierarchical stages with increasing feature dimensions and depthwise convolutions of large kernel size ($21\times21\times21$), followed by a symmetric decoder for volumetric segmentation. The encoder includes:

\begin{itemize}[leftmargin=*]
\item An initial input projection block with a $7\times7\times7$ convolution (stride 2, padding 3) followed by a residual block with two $3\times3\times3$ convolutions and GELU activations.
\item Stage 1: 2 Rep3D blocks with 48 channels followed by a strided $2\times2\times2$ convolution for downsampling.
\item Stage 2: 2 Rep3D blocks with 96 channels, followed by a strided $2\times2\times2$ convolution for downsampling.
\item Stage 3: 2 Rep3D blocks with 192 channels, followed by a strided $2\times2\times2$ convolution for downsampling.
\item Stage 4: 2 Rep3D blocks with 384 channels, followed by a strided $2\times2\times2$ convolution for downsampling.
\end{itemize}

Each stage modulates large kernel weights using a learnable re-parameterization mask computed via a lightweight 2-layer generator network within each Rep3D block. For each Rep3D block, it includes:
\begin{itemize}[leftmargin=*]
\item A single depthwise 3D convolution with a large kernel size of $21\times21\times21$ and padding size of $10$, followed by a layer normalization and a GELU activation.
\item A 2-stage lightweight generator network including:
\begin{itemize}
\item First layer: a depthwise $7\times7\times7$ convolution followed by layer normalization and a sigmoid activation.
\item Second layer: another depthwise $7\times7\times7$ convolution followed by layer normalization.
\end{itemize}
\end{itemize}

The decoder mirrors the encoder and consists of:

\begin{itemize}[leftmargin=*]
\item 4 upsampling modules (UnetrUpBlock from MONAI), each with a transpose convolution (stride 2), skip connection, and a residual block with two $3\times3\times3$ convolutions and GELU activations.
\item 1 output projection block (UnetOutBlock from MONAI) consisting of a $1\times1\times1$ convolution to map to the number of target classes.
\end{itemize}

\subsection{Training Efficiency Comparison}
To further validate our claims around improved convergence behavior and training efficiency, we conducted an additional ablation study focusing on runtime and performance dynamics across different configurations of Rep3D on the AMOS dataset. While our primary goal is to improve segmentation accuracy and spatial convergence, it is equally important that such gains are achieved with minimal training overhead. In this study, we compare three architectural variants:

\begin{enumerate}[label=(\alph*)]
    \item \textbf{Vanilla Rep:} Rep3D with parallel convolutional branches, but without any spatial prior modulation.
    \item \textbf{Fixed Prior:} Rep3D with a non-learnable reciprocal distance mask acting as a fixed spatial prior.
    \item \textbf{Full LRBM:} Rep3D with a learnable low-rank bias module (LRBM), modulating the spatial prior adaptively via a generator network.
\end{enumerate}

All models were trained under identical conditions: using a single NVIDIA A100 GPU, batch size of 2, and AdamW optimizer. We report the validation Dice scores at key training checkpoints (10k, 20k, 40k, and 60k iterations), as well as the total training time to convergence.

\begin{table}[H]
\centering
\caption{Training efficiency and convergence comparison of different Rep3D variants on AMOS dataset.}
\label{tab:training_efficiency}
\begin{tabular}{l|c|cccc}
\toprule
\textbf{Method} & \textbf{Time (hrs)} & \textbf{10k Iter} & \textbf{20k Iter} & \textbf{40k Iter} & \textbf{60k Iter} \\
\midrule
Vanilla Rep     & 17.3 & 0.853 & 0.868 & 0.886 & 0.892 \\
Fixed Prior     & 15.5 & 0.864 & 0.875 & 0.892 & 0.902 \\
Full LRBM       & 17.5 & \textbf{0.871} & \textbf{0.885} & \textbf{0.897} & \textbf{0.910} \\
\bottomrule
\end{tabular}
\end{table}

As observed in Table~\ref{tab:training_efficiency}, the full LRBM variant consistently achieves the best segmentation accuracy at every checkpoint, demonstrating accelerated convergence. The introduction of the learnable low-rank generator yields a modest increase in training time ($+0.2$ hours compared to Vanilla Rep), but this is substantially outweighed by the observed performance gains. The Fixed Prior variant also performs competitively, suggesting the benefit of incorporating even a static spatial prior.

\newpage
\subsection{Validation Experiments on Variable Branch Learning rate}
\begin{table*}[h]
    \centering
    \caption{Quantitative Evaluation on Variable Learning Rates in Parallel Branches}
    \begin{adjustbox}{width=0.85\textwidth}
    \begin{tabular}{*{1}{c}|*{2}{c}|*{3}{c}|*{1}{c}}
        \toprule
        Optimizer\: & \:Main Branch & Para. Branch  & \:Train Steps & Main LR & Para. LR\: & Mean Dice \\
        \midrule
        SGD & $21\times 21\times 21$ & $\times$ & 60000 & 0.0005 & $\times$ & 0.849\\
        SGD & $21\times 21\times 21$ & $\times$ & 60000 & 0.0004 & $\times$ & 0.852\\
        SGD & $21\times 21\times 21$ & $\times$ & 60000 & 0.0003 & $\times$ & 0.856\\
        SGD & $21\times 21\times 21$ & $\times$ & 60000 & 0.0002 & $\times$ & 0.859\\
        SGD & $21\times 21\times 21$ & $\times$ & 60000 & 0.0001 & $\times$ & 0.854\\
        \midrule
        AdamW & $21\times 21\times 21$ & $\times$ & 60000 & 0.0005 & $\times$ & 0.855\\
        AdamW & $21\times 21\times 21$ & $\times$ & 60000 & 0.0004 & $\times$ & 0.859\\
        AdamW & $21\times 21\times 21$ & $\times$ & 60000 & 0.0003 & $\times$ & 0.861\\
        AdamW & $21\times 21\times 21$ & $\times$ & 60000 & 0.0002 & $\times$ & 0.862\\
        AdamW & $21\times 21\times 21$ & $\times$ & 60000 & 0.0001 & $\times$ & 0.860\\
        \midrule
        SGD & $21\times 21\times 21$ & $3\times 3\times 3$ & 60000 & 0.0002 & 0.0006 & 0.872\\
        SGD & $21\times 21\times 21$ & $3\times 3\times 3$ & 60000 & 0.0002 & 0.0005 & 0.869\\
        SGD & $21\times 21\times 21$ & $3\times 3\times 3$ & 60000 & 0.0002 & 0.0004 & 0.867\\
        SGD & $21\times 21\times 21$ & $3\times 3\times 3$ & 60000 & 0.0002 & 0.0003 & 0.870\\
        SGD & $21\times 21\times 21$ & $3\times 3\times 3$ & 60000 & 0.0002 & 0.0001 & 0.865\\
        \midrule
        AdamW & $21\times 21\times 21$ & $3\times 3\times 3$ & 60000 & 0.0002 & 0.0006 & 0.887\\
        AdamW & $21\times 21\times 21$ & $3\times 3\times 3$ & 60000 & 0.0002 & 0.0005 & 0.886\\
        AdamW & $21\times 21\times 21$ & $3\times 3\times 3$ & 60000 & 0.0002 & 0.0004 & 0.887\\
        AdamW & $21\times 21\times 21$ & $3\times 3\times 3$ & 60000 & 0.0002 & 0.0003 & 0.889\\
        AdamW & $21\times 21\times 21$ & $3\times 3\times 3$ & 60000 & 0.0002 & 0.0001 & 0.886\\
        \bottomrule
    \end{tabular}
    \end{adjustbox}
\end{table*}

To empirically validate the theoretical insight of the spatially varying convergence dynamics in parallel-branched re-parameterization, we initially perform experiments using the CSLA block with Rep3D network architecture, composing of a main large kernel branch ($21\times21\times21$) and a parallel small kernel branch ($3\times3\times3$), with separate learning rates applied to each. As shown in Table 3, the single-branch design (no parallel branch) performance improved moderately with lower learning rates with both SGD and AdamW. The Dice score peaks at 0.859 with a learning rate of 0.0002 using SGD, and AdamW achieves its best performance of 0.862 at 0.0002 as well. However, with the addition of a small kernel parallel branch and using a higher learning rate for the small kernel (e.g., $\lambda_{\text{S}} > \lambda_{\text{L}}$), we observed consistent improvements across all configurations. Specifically, the best result with SGD reached 0.872 when using $\lambda_{L} = 0.0002$ and $\lambda_{S} = 0.0006$. Similarly, AdamW attained a maximum Dice score of 0.889 with $\lambda_{L} = 0.0002$ and $\lambda_{S} = 0.0003$. These results validate our hypothesis that assigning higher learning rates to the small kernel branch accelerates convergence of central kernel regions, while maintaining stability in peripheral regions with a lower learning rate for the large kernel. Moreover, such results further confirm that spatially varying convergence behavior can be approximated through differentiated learning rates, supporting the design principle behind our learnable re-parameterization in Rep3D.

\section{Ablation Study on Network Depth for LRBM}
\begin{table*}[h]
    \centering
    \caption{Ablation Study on Network Depth for LRBM with the AMOS testing split}
    \begin{adjustbox}{width=\textwidth}
    \begin{tabular}{l|ccccccccccccccc|c}
        \toprule
        Number of Layers & \multicolumn{1}{c}{Spleen} & \multicolumn{1}{c}{R. Kid} & \multicolumn{1}{c}{L. Kid} & \multicolumn{1}{c}{Gall.} & \multicolumn{1}{c}{Eso.} & \multicolumn{1}{c}{Liver} & \multicolumn{1}{c}{Stom.} & \multicolumn{1}{c}{Aorta} & \multicolumn{1}{c}{IVC} & \multicolumn{1}{c}{Panc.} & \multicolumn{1}{c}{RAG} & \multicolumn{1}{c}{LAG} & \multicolumn{1}{c}{Duo.} & \multicolumn{1}{c}{Blad.} & \multicolumn{1}{c|}{Pros.} & \multicolumn{1}{c}{Avg}\\ 
        \midrule
        1 Layer & 0.974 & 0.965 & 0.964 & 0.925 & 0.859 & 0.982 & 0.926 & 0.956 & 0.920 & 0.842 & 0.824 & 0.781 & 0.842 & 0.915 & 0.879 & 0.904 \\
        2 Layers & \textbf{0.978} & \textbf{0.970} & 0.964 & \textbf{0.928} & \textbf{0.871} & \textbf{0.984} & \textbf{0.927} & \textbf{0.960} & \textbf{0.930} & \textbf{0.851} & \textbf{0.828} & \textbf{0.784} & \textbf{0.850} & \textbf{0.920} & \textbf{0.881} & \textbf{0.910} \\
        3 Layers & 0.971 & 0.964 & \textbf{0.965} & 0.924 & 0.841 & 0.983 & 0.920 & 0.952 & 0.910 & 0.839 & 0.819 & 0.779 & 0.837 & 0.910 & 0.870 & 0.899 \\
        \bottomrule
    \end{tabular}
    \end{adjustbox}
\end{table*}

\subsection{Additional Comparisons with nnU-Net Variants: ResEnc nnU-Net, STU-Net, MedNeXt}
we conducted further comparisons with recent state-of-the-art (SOTA) 3D medical image segmentation architectures, including \textbf{ResEnc nnU-Net}~\cite{isensee2021nnu}, \textbf{STU-Net-H}~\cite{huang2023stu}, and \textbf{MedNeXt}~\cite{roy2023mednext}. These models have demonstrated strong performance across various benchmarks and provide important context for positioning Rep3D among contemporary architectures. All baseline methods were trained under their official recommended training schedules (typically 1000 epochs), using the same computational resources and training splits for fair comparison.

\begin{table}[h]
\centering
\caption{Average Dice scores across five segmentation benchmarks under full training schedules.}
\begin{tabular}{l|ccccc}
\toprule
\textbf{Method} & \textbf{AMOS CT} & \textbf{AMOS MRI} & \textbf{KiTS} & \textbf{Pancreas} & \textbf{Hepatic} \\
\midrule
nnU-Net (1000 epochs) & 0.887 & 0.847 & 0.706 & 0.703 & 0.660 \\
ResEnc nnU-Net & 0.892 & 0.850 & 0.711 & 0.706 & 0.661 \\
STU-Net-H & 0.900 & 0.848 & 0.707 & 0.712 & 0.648  \\
MedNeXt & 0.897 & 0.856 & 0.720 & 0.713 & 0.663  \\
\textbf{Rep3D (Ours)} & \textbf{0.910} & \textbf{0.864} & \textbf{0.736} & \textbf{0.723} & \textbf{0.674} \\
\bottomrule
\end{tabular}
\label{tab:combined_results}
\end{table}

\noindent These comparisons further validate the strong and consistent performance of \textbf{Rep3D} across multiple challenging 3D segmentation benchmarks. Even under extensive training schedules (1000 epochs), Rep3D outperforms the SOTA alternatives, demonstrating its robustness and generalizability.




\end{document}